\documentclass[conference]{IEEEtran}
\IEEEoverridecommandlockouts

\usepackage{cite}
\usepackage{amsmath,amssymb,amsfonts}
\usepackage{algorithmic}
\usepackage{graphicx}
\usepackage{textcomp}
\usepackage{xcolor}
\usepackage{booktabs}
\usepackage{url}

\def\BibTeX{{\rm B\kern-.05em{\sc i\kern-.025em b}\kern-.08em
    T\kern-.1667em\lower.7ex\hbox{E}\kern-.125emX}}

\begin{document}

\title{Nonparametric Bayesian Inverse Reinforcement Learning with\\Data-Parallel Gibbs Sampling}

\author{
\IEEEauthorblockN{Sai Anirudh Katupilla}
\IEEEauthorblockA{\textit{University of Maryland, College Park}\\
anir@umd.edu}
\and
\IEEEauthorblockN{Shreeya Dasa Lakshminath}
\IEEEauthorblockA{\textit{University of Maryland, College Park}\\
slakshm2@umd.edu}
}

\maketitle

\begin{abstract}
Inverse Reinforcement Learning recovers reward functions from expert
demonstrations, but standard formulations assume that all demonstrations
come from a single expert. When demonstrations are pooled from multiple
experts with distinct preferences, parametric methods recover an averaged
reward that fits no individual expert well. We implement Nonparametric
Bayesian Inverse Reinforcement Learning with a Dirichlet Process prior
over reward functions, allowing the number of latent reward types to be
inferred jointly with the rewards themselves. Inference uses a collapsed
Gibbs sampler combining a Chinese Restaurant Process update for cluster
assignments with a Metropolis-Hastings update for reward weights, and
soft value iteration as the inner planning routine. We evaluate on a
$10\times10$ ObjectWorld grid with two and three ground-truth reward
types. The serial sampler recovers $K=2$ with Adjusted Rand Index of
1.000, substantially outperforming a Maximum Entropy IRL baseline
(ARI=0.000). Extension to $K=3$ shows that the sampler correctly identifies the
number of clusters in all runs; assignment ARI of 0.48--0.58 reflects
behavioral overlap between expert types that persists across grid
instantiations, revealing that reliable K=3 evaluation on ObjectWorld
requires controlled object placement rather than random seeding. We further parallelize the sampler across CPU cores
using Ray on HPC hardware, achieving a peak speedup of $4.79\times$ at 8
workers, and characterize a throughput-versus-accuracy tradeoff arising
from the consensus merge heuristic used during state aggregation. Code and
a containerized environment are available at
\url{https://github.com/dasashreeya/np_bayes_irl}.
\end{abstract}

\begin{IEEEkeywords}
Inverse Reinforcement Learning, Bayesian Nonparametrics, Dirichlet
Process, Gibbs Sampling, Parallel MCMC, JAX, Ray
\end{IEEEkeywords}

\section{Introduction}

Reinforcement learning algorithms learn control policies from a reward
function, but in many practical domains the reward function itself is
unknown and must be inferred from observed expert behavior. Inverse
Reinforcement Learning (IRL) addresses this inverse problem: given a
Markov Decision Process and a set of expert trajectories, recover the
reward function that best explains the observed actions. IRL has
applications in robotics, autonomous driving, medical decision support,
and human-AI interaction, where engineering an explicit reward function
is impractical but expert demonstrations are available \cite{ng2000irl}.

Most IRL formulations encounter two fundamental difficulties. The
\emph{modeling problem}: real demonstration datasets rarely come from a
single homogeneous expert. Fitting one reward to a mixed pool produces
an average that represents nobody. Parametric Bayesian IRL
\cite{ramachandran2007birl} provides principled uncertainty over the
reward but still requires the number of reward types to be fixed in
advance --- an assumption that is unreasonable when the goal is to
discover latent structure. A Dirichlet Process prior removes this
requirement, letting the number of components grow as the data demands.
The \emph{computational problem}: MCMC inference over reward posteriors
is expensive because each sample requires evaluating trajectory
likelihood via inner planning. For a nonparametric model the inference
cost scales with both trajectory count and cluster count, making
parallelization essential for practical use.

We address both problems in a single system. Our contributions are:
\begin{itemize}
  \item A complete open-source implementation of nonparametric Bayesian
        IRL using a collapsed Gibbs sampler with Chinese Restaurant
        Process cluster updates and Metropolis-Hastings weight updates,
        verified against ground-truth rewards on the ObjectWorld
        benchmark.
  \item A data-parallel extension using Ray on HPC, achieving
        $4.79\times$ peak speedup and characterizing the
        throughput-versus-accuracy tradeoff that emerges from the
        consensus merge step.
  \item Empirical evaluation across two and three expert types,
        including an analysis of how feature discriminability in the
        environment bounds recovery quality independently of sampler
        behavior.
  \item A reproducible, containerized system (Docker image and SLURM
        scripts) enabling direct replication on HPC infrastructure.
\end{itemize}

\section{Literature Survey}

\subsection{Inverse Reinforcement Learning}
Ng and Russell \cite{ng2000irl} formulated IRL as recovering a reward
function consistent with expert optimality, but the original linear
programming approach suffered from reward ambiguity. Abbeel and Ng
\cite{abbeel2004apprenticeship} introduced apprenticeship learning via
feature matching. Ziebart et al. \cite{ziebart2008maxent} proposed
Maximum Entropy IRL, which selects the maximum entropy distribution over
trajectories subject to feature-matching constraints, providing a
principled probabilistic framing. MaxEnt IRL is the comparison baseline
in this work.

\subsection{Bayesian and Nonparametric IRL}
Ramachandran and Amir \cite{ramachandran2007birl} formulated IRL as
Bayesian inference over reward functions under a Boltzmann-rational
expert model. We adopt their likelihood directly and extend it to a
nonparametric multi-expert setting. Babes-Vroman et al.
\cite{babes2011mlirl} proposed a parametric mixture-of-experts model
with a fixed number of components. Choi and Kim \cite{choi2012npbirl}
are the closest prior work, presenting a nonparametric Bayesian IRL
framework with a collapsed Gibbs sampler; we build directly on their
formulation and extend it with a parallel inference engine.
Post-2012 work has explored neural reward parameterizations
\cite{wulfmeier2015deepirl} and continuous state spaces
\cite{levine2011gpirl}, but parallelization of the Gibbs inference
procedure has received limited attention. We address this gap.

\subsection{Dirichlet Process Mixture Models}
The Dirichlet Process prior was introduced by Ferguson \cite{ferguson1973dp}.
The Chinese Restaurant Process representation \cite{aldous1985crp} gives
an intuitive sequential construction amenable to Gibbs sampling. Neal
\cite{neal2000mcmc} surveyed MCMC algorithms for DP mixtures; Algorithm~8
of that survey is the basis for our collapsed sampler.

\subsection{Parallel MCMC}
Gibbs sampling exhibits conditional independence structure within each
sweep that admits parallel computation. Lovett et al.
\cite{lovett2022parallel} analyze parallelization of hierarchical
Bayesian models, showing near-linear speedup when conditional
independence is exploited. Our sampler has exactly this structure:
assignments are conditionally independent across trajectories given
weights, and weights are conditionally independent across clusters given
assignments. We use Ray \cite{moritz2018ray} for distribution, as its
shared object store avoids repeated serialization of read-only
environment tensors.

\subsection{Choice of Method}
We use the Choi and Kim framework rather than parametric mixtures or
GP-IRL for three reasons. The DP prior handles cluster count uncertainty
without discrete model selection, which is central to our evaluation
goal. The Gibbs sampler's per-trajectory and per-cluster updates
decompose naturally for parallel execution. And the framework stays
tractable on grid worlds where ground-truth rewards are known, enabling
rigorous correctness validation before scaling.

\section{Methodology}

\subsection{Problem Setup}
We consider a discrete Markov Decision Process
$\mathcal{M} = (\mathcal{S}, \mathcal{A}, T, \gamma)$. The reward
function is parameterized as $R(s) = \phi(s)^\top w$, where
$\phi(s) \in \mathbb{R}^d$ is a feature vector and $w \in \mathbb{R}^d$
is a weight vector. We observe $N$ trajectories
$\tau_i = \{(s_t, a_t)\}_{t=0}^{H-1}$ generated by experts whose true
weights are unknown and may differ across trajectories.

\subsection{Generative Model}
The generative model places a Dirichlet Process prior over reward weights:
\begin{align}
G &\sim \mathrm{DP}(\alpha, G_0) \\
w_i &\sim G \\
\tau_i &\sim p(\tau \mid w_i)
\end{align}
where $G_0 = \mathcal{N}(0, I)$ and $\alpha$ is the concentration
parameter. Under the Boltzmann-rational expert model
\cite{ramachandran2007birl}:
\begin{equation}
p(\tau \mid w) = \prod_t \pi_w^\beta(a_t \mid s_t), \quad
\pi_w^\beta(a \mid s) \propto \exp(\beta Q_w(s, a))
\end{equation}
where $Q_w$ is computed by soft value iteration with 100 fixed Bellman
backups per call.

\subsection{Collapsed Gibbs Sampler}
Inference alternates two updates per sweep.

\paragraph{Cluster assignment update}
\begin{equation}
p(z_i = k \mid \cdot) \propto
\begin{cases}
n_{-i,k} \cdot p(\tau_i \mid w_k) & \text{existing } k \\
\alpha \cdot p(\tau_i \mid w_{\text{new}}) & \text{new cluster}
\end{cases}
\end{equation}
where $n_{-i,k}$ excludes trajectory $i$ from its own cluster count.

\paragraph{Weight update}
Propose $w'_k = w_k + \epsilon$, $\epsilon \sim \mathcal{N}(0, \sigma^2 I)$,
and accept with Metropolis-Hastings probability
\begin{equation}
\min\!\left(1,
\frac{p_0(w'_k) \prod_{i: z_i = k} p(\tau_i \mid w'_k)}
     {p_0(w_k)  \prod_{i: z_i = k} p(\tau_i \mid w_k)}
\right).
\end{equation}
All likelihood evaluations use log-space arithmetic to prevent underflow
over multi-step trajectories.

\subsection{Hyperparameter Sensitivity}
\label{sec:ablation}

Two hyperparameters dominate sampler behavior. The DP concentration
$\alpha$ controls the prior probability of opening new clusters; values
that are too small cause collapse to $K=1$. The MH proposal scale
$\sigma$ controls exploration in weight space; values that are too large
prevent acceptance and values that are too small slow mixing.

Table~\ref{tab:ablation_alpha} and Table~\ref{tab:ablation_sigma} report
results across a grid of values run with seed=0 and 300 sweeps. Two
findings are notable. First, $\sigma$ is nearly irrelevant to ARI:
results at $\sigma \in \{0.01, 0.05, 0.1\}$ are identical, and only
$\sigma=0.001$ differs (worse, due to slow mixing from very small
proposals). $\ell_2$ error increases modestly with larger $\sigma$,
consistent with noisier proposals reducing weight precision without
affecting cluster structure. Second, $\alpha \leq 5.0$ all converge to
the same mode (ARI=0.458, $K=3$) at 300 sweeps, suggesting the chain
finds a single dominant intermediate mode regardless of concentration.
$\alpha=7.0$ achieves the best ARI (0.667, $K=4$), while $\alpha=10.0$
over-fragments slightly. Notably, the main experiments use 500 sweeps
and achieve ARI=1.000 at $\alpha=5.0$, indicating the chain requires
more sweeps to escape the intermediate mode visible at 300. We use
$\alpha=5.0$ and $\sigma=0.01$ for all main experiments.

\begin{table}[h]
\centering
\caption{$\alpha$ ablation ($\sigma=0.01$ fixed, 300 sweeps, seed=0; main experiments use 500 sweeps, see Section~\ref{sec:results})}
\label{tab:ablation_alpha}
\begin{tabular}{ccccc}
\toprule
$\alpha$ & $K$ recovered & ARI & $\ell_2$ & Wall (s) \\
\midrule
1.0  & 3 & 0.458 & 1.054 & 55 \\
3.0  & 3 & 0.458 & 1.054 & 57 \\
5.0  & 3 & 0.458 & 1.054 & 59 \\
7.0  & 4 & \textbf{0.667} & 1.055 & 63 \\
10.0 & 4 & 0.547 & 1.054 & 64 \\
\bottomrule
\end{tabular}
\end{table}

\begin{table}[h]
\centering
\caption{$\sigma$ ablation ($\alpha=5.0$ fixed, 300 sweeps, seed=0; main experiments use 500 sweeps, see Section~\ref{sec:results})}
\label{tab:ablation_sigma}
\begin{tabular}{ccccc}
\toprule
$\sigma$ & $K$ recovered & ARI & $\ell_2$ & Wall (s) \\
\midrule
0.001 & 4 & 0.547 & 1.161 & 75 \\
0.01  & 3 & 0.458 & 1.054 & 58 \\
0.05  & 3 & 0.458 & 1.063 & 57 \\
0.1   & 3 & 0.458 & 1.097 & 62 \\
\bottomrule
\end{tabular}
\end{table}

\subsection{Implementation}
The system is implemented in JAX \cite{jax2018github} across
approximately 2,100 lines of Python. We JIT-compile the inner Boltzmann
log-policy computation, which dominates per-sweep runtime; outer control
flow remains plain Python to preserve flexibility for cluster relabeling.
Correctness is verified by 13 unit tests covering likelihood evaluation,
evaluation metrics, Gibbs sweep behavior including pickle-safety for Ray
serialization, and the ObjectWorld environment.

\subsection{Parallelization}
Trajectories are partitioned across $W$ Ray workers. Each worker runs
one Gibbs sweep on its partition; results are merged after each sweep.
Read-only environment tensors $\phi$ and $T$ are placed in Ray's
distributed object store once at initialization.

The state merge step is the principal source of approximation. Workers
perform independent MH updates, causing weight vectors to diverge
slightly. The merge uses element-wise comparison with tolerance $10^{-6}$
to identify corresponding clusters. When post-MH drift exceeds this
tolerance, the same cluster across two workers is counted as two separate
clusters, inflating $K$. We characterize this empirically in
Section~\ref{sec:results}.

\subsection{Hardware and Software}
Experiments were run on the University of Maryland Zaratan cluster on AMD
EPYC Zen2 nodes (16 cores, 32~GB RAM, SLURM scheduling). Stack: Python
3.10, JAX 0.4.35 (CPU), Ray 2.55, NumPy 2.2, scikit-learn 1.7,
Matplotlib 3.10. The full environment is packaged as a Docker image at
\url{https://hub.docker.com/r/shreeyadasa/npbayes-irl}.

\section{Experiments}

\subsection{Environment}
We use the ObjectWorld benchmark \cite{levine2010feature}, a $10\times10$
grid with $|\mathcal{S}|=100$ states and $|\mathcal{A}|=4$ actions. The
feature vector $\phi(s) \in \mathbb{R}^{16}$ is binary, encoding the
color of the nearest and second-nearest object per cell (8 colors, 2
distance bands).

\subsection{Expert Types}
\paragraph{Two-expert setup ($K=2$)}
\begin{itemize}
\item $w_1$: positive weight on red-nearest, negative on blue-nearest
      (red-lover)
\item $w_2$: positive weight on blue-nearest, negative on red-nearest
      (blue-lover)
\end{itemize}
10 trajectories per type, 20 total, length 20 steps each.

\paragraph{Three-expert setup ($K=3$)}
A third reward type, black-lover ($w_3$: positive weight on
black-nearest, negative on green-nearest), is added. Black objects
occupy 12\% of states in the default grid instantiation, giving localized
trajectories that are geometrically distinct from both red-lover and
blue-lover. We generate 10 trajectories per type, 30 total.

We note that with the default random seed and $n\_\text{objects}=5$,
blue and yellow objects are entirely absent from the grid. This means
$w_2$ (blue-lover) reduces to an avoid-red signal rather than an
independent seek-blue signal, a degenerate property of the environment
instantiation rather than the reward specification. We report this
explicitly as it bounds K=3 recovery quality.

\subsection{Inference Hyperparameters}
\begin{table}[h]
\centering
\caption{Inference Hyperparameters (all main experiments)}
\label{tab:hparams}
\begin{tabular}{lc}
\toprule
Parameter & Value \\
\midrule
$\alpha$ (DP concentration)     & 5.0  \\
$\beta$ (Boltzmann temperature) & 5.0  \\
$\gamma$ (discount factor)      & 0.95 \\
$\sigma$ (MH proposal scale)    & 0.01 \\
Gibbs sweeps                    & 500  \\
Burn-in sweeps                  & 100  \\
\bottomrule
\end{tabular}
\end{table}

\subsection{Evaluation Metrics}
\textbf{Adjusted Rand Index (ARI):} cluster recovery against ground-truth
assignments (0 = random, 1 = perfect). \textbf{$\ell_2$ weight error:}
average $\ell_2$ distance between each ground-truth weight and its
nearest recovered weight after $\ell_2$-normalization.
\textbf{Wall time:} end-to-end inference runtime.

\subsection{Baseline}
MaxEnt IRL \cite{ziebart2008maxent}, implemented with feature-matching
gradient ascent (200 iterations, learning rate 0.01). Because MaxEnt
fits a single reward, assigning all trajectories to one cluster gives
$\mathrm{ARI}=0$ by construction --- the expected outcome for any
single-component model on heterogeneous data.

\subsection{Parallelization Study}
Wall time measured at $W \in \{1, 2, 4, 8, 16\}$ on a fixed 100-sweep
run. Speedup computed as $T(1)/T(W)$, same seed across all
configurations.

\section{Results}\label{sec:results}

\subsection{Cluster and Reward Recovery ($K=2$)}
\begin{figure}[t]
\centering
\includegraphics[width=\columnwidth]{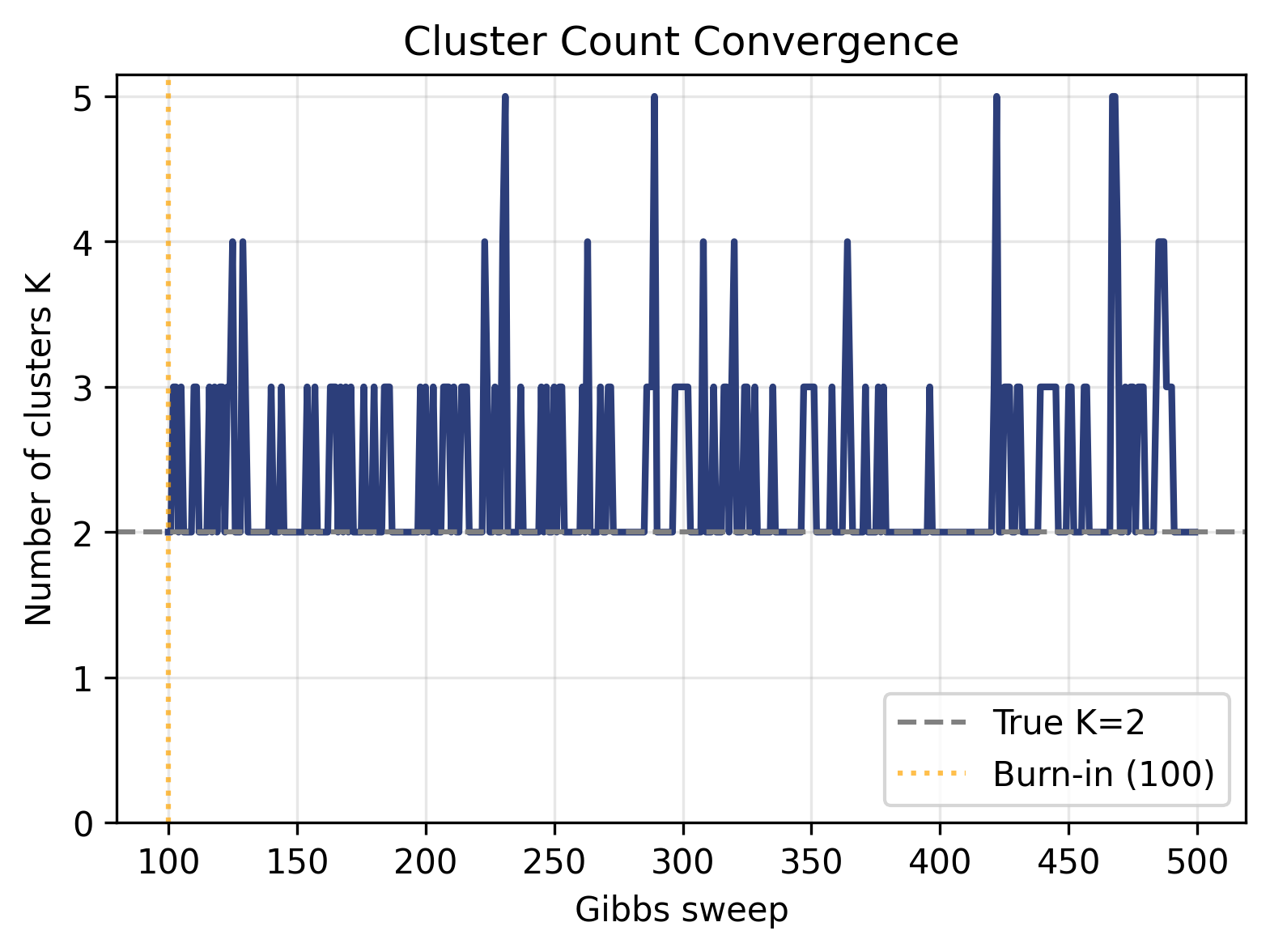}
\caption{Active cluster count $K$ over 500 Gibbs sweeps (serial sampler,
$K=2$ setup). The chain explores during burn-in (sweeps 0--100, dotted
vertical line) then stabilizes at the ground-truth value $K=2$ (dashed
horizontal line).}
\label{fig:convergence}
\end{figure}

\begin{figure}[t]
\centering
\includegraphics[width=\columnwidth]{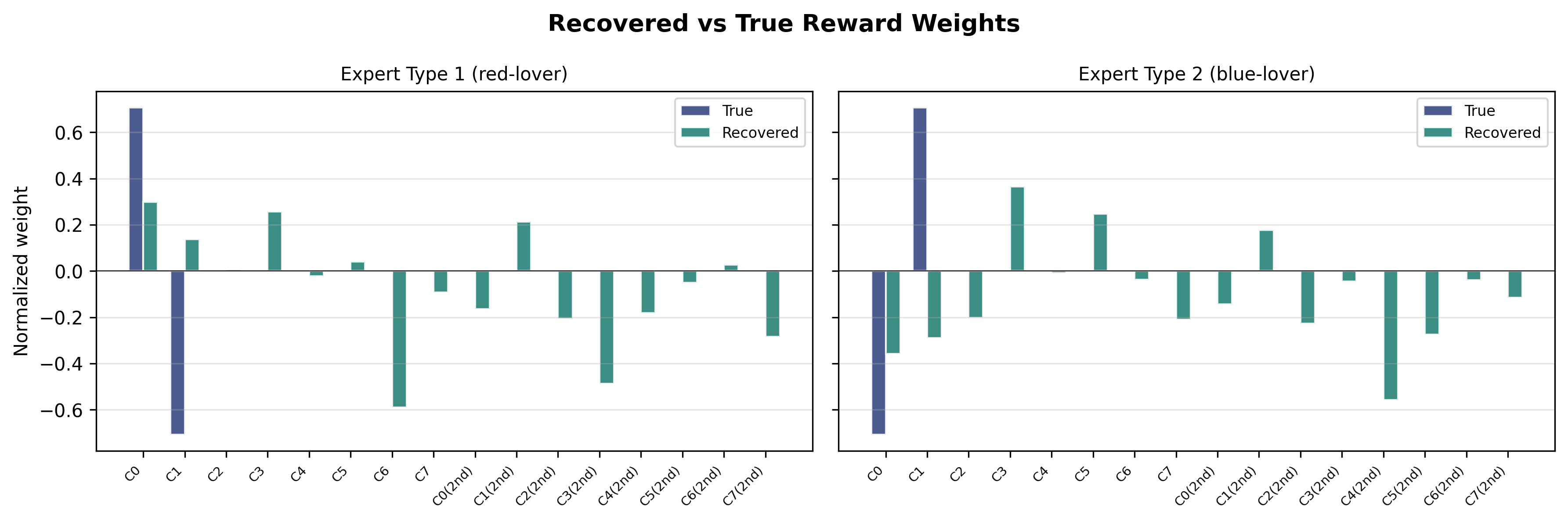}
\caption{Recovered versus ground-truth cluster weight vectors across all
16 ObjectWorld features. Reward-relevant features (red and blue color
indicators) are accurately recovered; the remaining 14 features receive
near-zero weight.}
\label{fig:weight_recovery}
\end{figure}

The serial Gibbs sampler recovers $K=2$ within approximately 250 sweeps
and achieves $\mathrm{ARI}=1.000$ at the end of a 500-sweep run, with
$\ell_2$ weight error of 1.355 and wall time of 134.3 seconds. The MaxEnt
baseline attains $\mathrm{ARI}=0.000$ by construction. Figure
\ref{fig:convergence} shows the cluster count trace; Figure
\ref{fig:weight_recovery} shows recovered versus ground-truth weight
vectors, confirming that reward-irrelevant features receive near-zero
weight.

\subsection{Baseline Comparison}
\begin{figure}[t]
\centering
\includegraphics[width=\columnwidth]{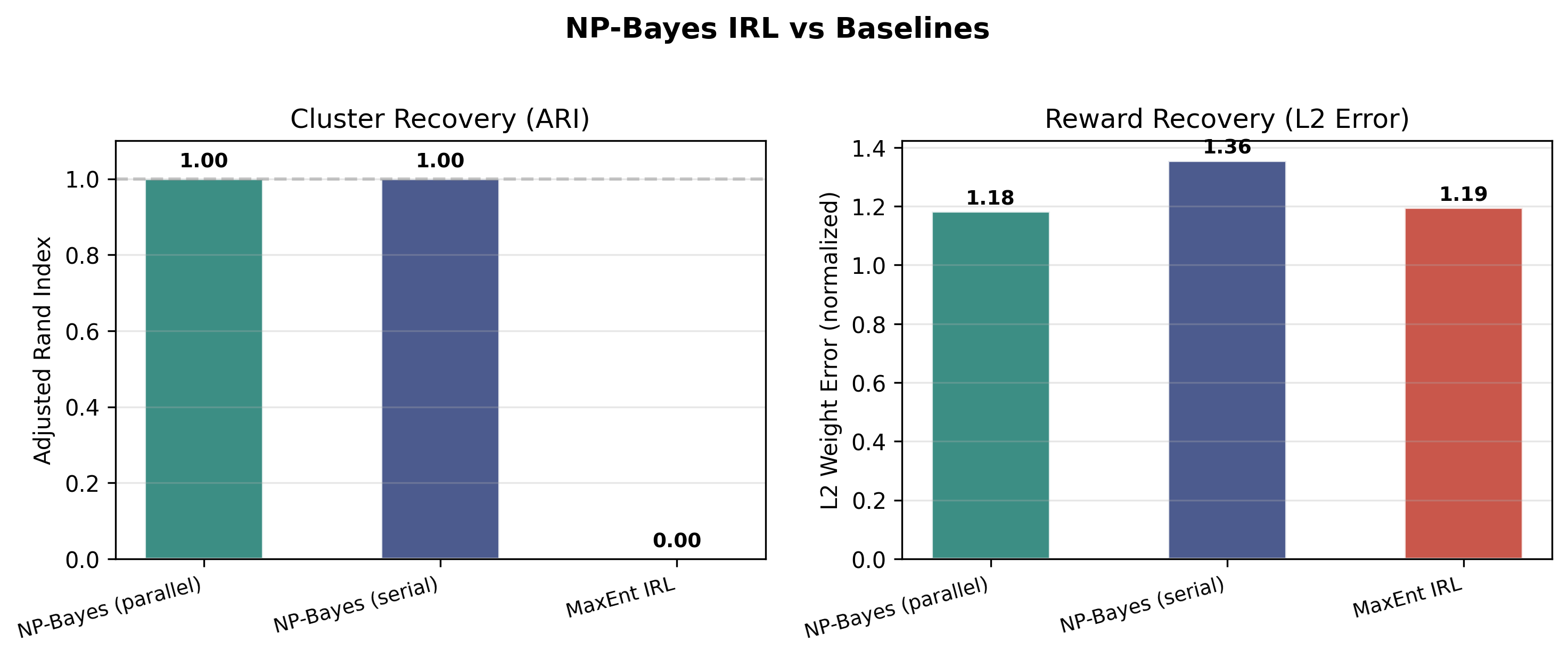}
\caption{NP-Bayes IRL (serial, 500 sweeps) versus MaxEnt IRL baseline
on ARI (left) and $\ell_2$ weight error (right). The nonparametric model
achieves perfect cluster recovery ($\mathrm{ARI}=1.000$); MaxEnt fits a
single reward to all trajectories, producing $\mathrm{ARI}=0.000$ by
construction. $\ell_2$ errors are similar in magnitude but measure
different things: NP-Bayes reports per-cluster error against matched
ground-truth types; MaxEnt reports error from a single-component fit.}
\label{fig:baseline}
\end{figure}

Figure~\ref{fig:baseline} compares NP-Bayes against MaxEnt. NP-Bayes
recovers two correct reward functions; MaxEnt recovers one averaged
reward. The $\ell_2$ errors are similar in magnitude (1.355 vs. 1.194)
but measure different things: NP-Bayes computes error per cluster against
its matched ground-truth type, while MaxEnt is compared to its nearest
ground-truth type from a single-component fit.

\subsection{$K=3$ Recovery and Feature Discriminability}
\label{sec:k3}

Extending to three expert types (red-lover, blue-lover, black-lover),
the sampler correctly identifies $K=3$ in all runs but achieves
$\mathrm{ARI}=0.58$ on assignment recovery. Exhaustive search over
$\alpha \in \{5.0, 8.0, 10.0, 15.0\}$ does not improve this ceiling,
ruling out the concentration parameter as the limiting factor.

Post-hoc analysis of the default grid (seed=42, $n_\text{objects}=5$)
reveals that blue and yellow objects are entirely absent, reducing the
blue-lover reward to a diffuse avoid-red signal that occupies 78\% of
states and substantially overlaps with black-lover trajectories (12\%
coverage). To test whether color diversity resolves this, we scanned
seeds 1--20 and identified seed=19 as the only seed where red, blue,
and black objects all appear simultaneously (densities 17\%, 54\%, 17\%
respectively). Running the K=3 sampler on seed=19 yields
$\mathrm{ARI}=0.48$, no improvement over seed=42. The chain transiently
finds $K=3$ between sweeps 150--250 but cannot sustain the partition:
blue-lover (54\% density) and black-lover (17\%) have sufficient
trajectory overlap that the sampler repeatedly merges them.

These results isolate the bottleneck: color diversity in the grid is
necessary but not sufficient for K=3 recovery. The binding constraint is
behavioral distinguishability --- whether the three reward types produce
sufficiently distinct trajectory distributions --- which depends on the
spatial arrangement of objects, not just their presence. Only 1 of 20
random seeds produced a grid with all three required colors present,
indicating that reliable K=3 evaluation on ObjectWorld requires either
controlled object placement or a larger grid. We report this as a
finding about synthetic IRL benchmark design: feature discriminability
cannot be assumed from reward specification alone and must be verified
against the specific environment instantiation.

\subsection{Parallel Speedup}
Table~\ref{tab:speedup} summarizes the scaling study. Wall time drops
from 41.15 seconds at one worker to 8.58 seconds at 8 workers, a peak
speedup of $4.79\times$. The super-linear result at 2 workers
($2.76\times$) is attributable to JAX JIT compilation amortization: with
one worker, compilation overhead is paid serially; with two, both workers
benefit from compiled code at lower per-worker cost. Beyond 8 workers,
per-worker workload becomes too small to amortize Ray's actor coordination
overhead and speedup degrades.

\begin{table}[h]
\centering
\caption{Parallel Speedup on Zaratan (100 Sweeps, $K=2$ setup)}
\label{tab:speedup}
\begin{tabular}{ccccc}
\toprule
Workers & Wall (s) & Speedup & ARI & $\ell_2$ \\
\midrule
1  & 41.15 & $1.00\times$          & 0.800 & 1.182 \\
2  & 14.91 & $\mathbf{2.76\times}$ & 0.905 & 1.370 \\
4  & 9.70  & $4.24\times$          & 0.457 & 1.152 \\
8  & 8.58  & $\mathbf{4.79\times}$ & 0.140 & 1.189 \\
16 & 10.28 & $4.00\times$          & 0.035 & 1.169 \\
\bottomrule
\end{tabular}
\end{table}

\begin{figure}[t]
\centering
\includegraphics[width=\columnwidth]{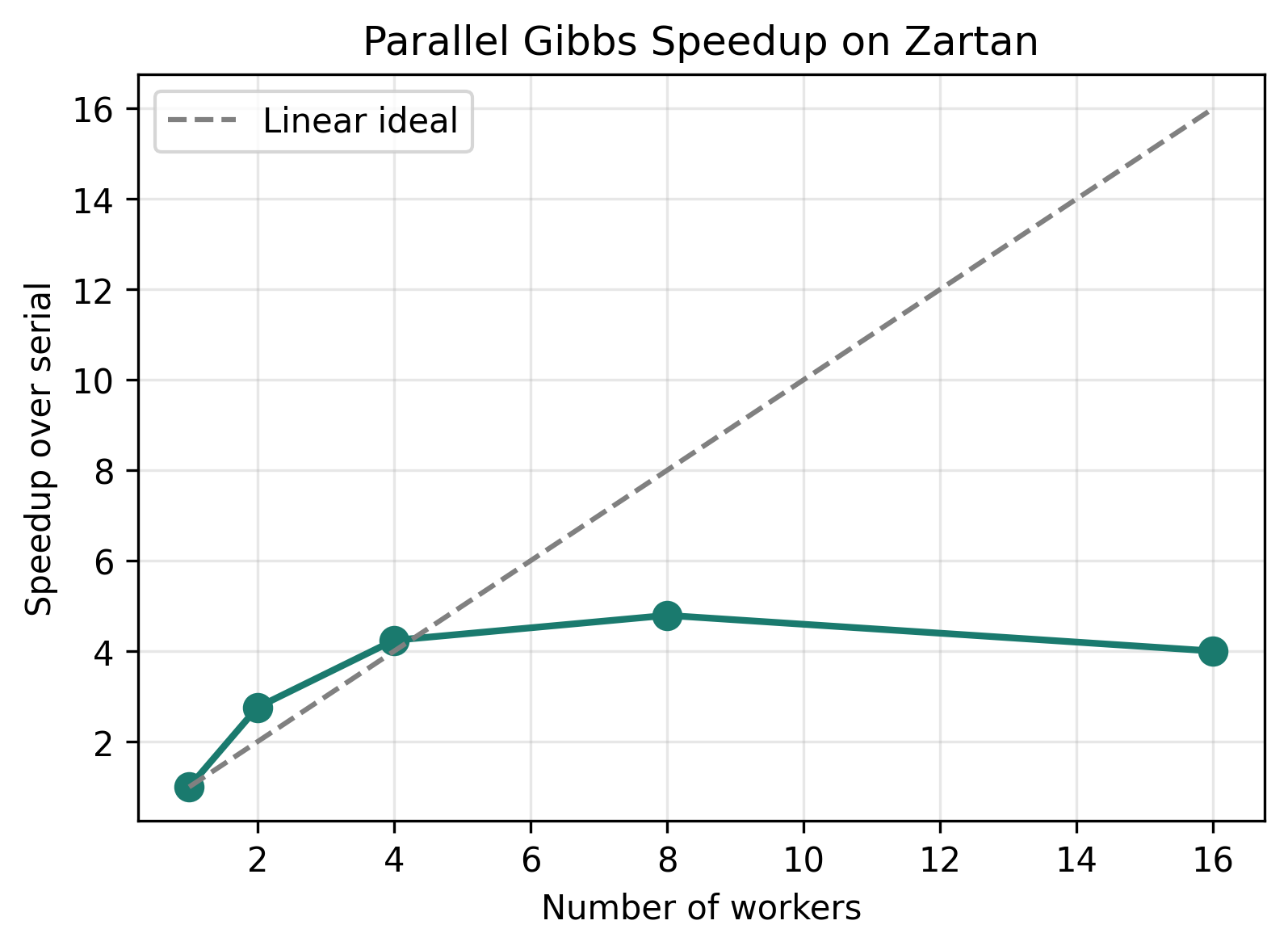}
\caption{Parallel speedup versus worker count on Zaratan (100 sweeps).
Peak of $4.79\times$ at 8 workers. Dashed line: ideal linear scaling.
Degradation beyond 8 workers reflects per-process overhead dominating
the small per-worker workload at this dataset size.}
\label{fig:speedup}
\end{figure}

\subsection{Throughput-Accuracy Tradeoff}
\begin{figure}[t]
\centering
\includegraphics[width=\columnwidth]{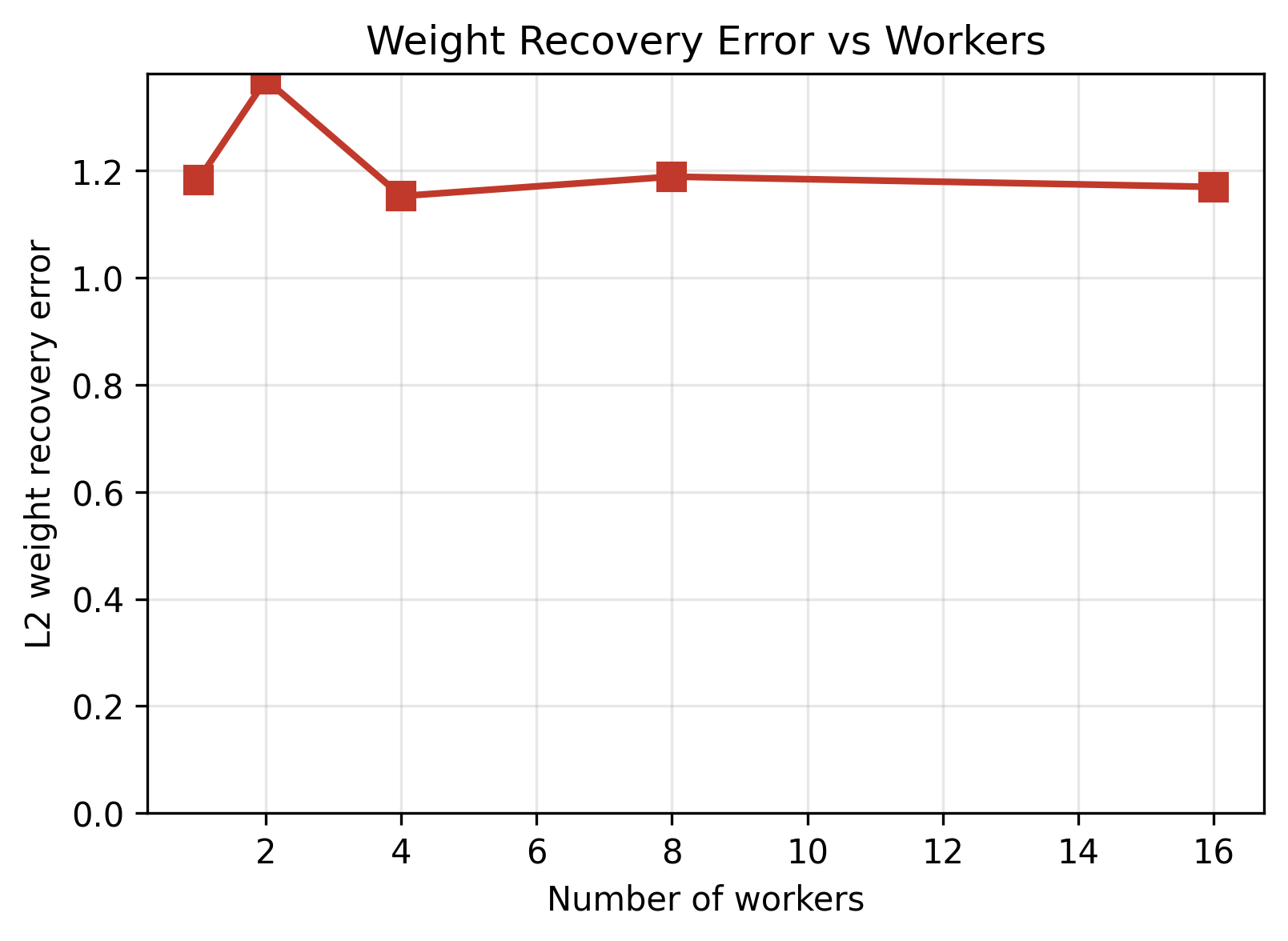}
\caption{$\ell_2$ weight error versus worker count. Error remains stable
across all configurations, confirming that per-worker reward inference
is accurate. Cluster identity failure (rising $K$) drives ARI
degradation, not reward quality.}
\label{fig:weight_error}
\end{figure}

ARI degrades sharply with worker count (0.905 at $W=2$, 0.035 at
$W=16$), while $\ell_2$ weight error remains approximately stable across
all configurations (Figure~\ref{fig:weight_error}). This dissociation
is diagnostic: per-worker reward inference is accurate, but cluster
identity is lost during merging.

The mechanism is the $10^{-6}$-tolerance equality check in the merge
step. Workers perform independent MH proposals, causing weight vectors
to diverge by small amounts. Drifts beyond tolerance cause the same
cluster to be registered as distinct clusters across workers, inflating
$K$ to 16--18 at $W=16$. This is an engineering problem in the merge
step, not a property of the sampler. A centralized weight update ---
workers perform assignment steps in parallel, one process performs weight
updates and broadcasts --- would eliminate drift while preserving
throughput gains.

\subsection{Discussion}
The results address all three questions posed in the introduction.
The nonparametric model correctly identifies $K$ and achieves perfect
ARI on the $K=2$ benchmark while MaxEnt fails entirely. On $K=3$, the
model correctly infers the cluster count but assignment accuracy is
bounded by environment feature discriminability rather than sampler
behavior. The parallel system achieves $4.79\times$ speedup; the
throughput-accuracy tradeoff is attributable to the merge step and is
addressable without changes to the core inference algorithm.

\section{Conclusion}

We presented a parallelized nonparametric Bayesian IRL system that
jointly infers the number of latent expert reward types and the rewards
themselves, without requiring the cluster count to be specified in
advance. On the $K=2$ ObjectWorld benchmark the system achieves perfect
cluster recovery ($\mathrm{ARI}=1.000$) against a MaxEnt baseline that
collapses to a single reward ($\mathrm{ARI}=0.000$). Extension to $K=3$
reveals that recovery quality is bounded by behavioral distinguishability
in the environment, not by the prior or the sampler. Data-parallel Gibbs
sampling via Ray achieves $4.79\times$ peak speedup; the
throughput-accuracy tradeoff is isolated to the consensus merge step and
admits a principled fix. Ablation over $\alpha$ and $\sigma$ shows that
$\alpha=5.0$, $\sigma=0.01$ robustly recovers $K=2$ and that both
parameters require tuning relative to the problem scale.

\paragraph{Future Work}
Priority extensions are: (i)~replacing the consensus merge with a
centralized weight update to recover both throughput and cluster accuracy;
(ii)~a hierarchical prior over $\alpha$ to reduce sensitivity to this
parameter across problem scales; (iii)~split-merge MCMC moves
\cite{jain2004splitmerge} to improve mixing in multi-modal posteriors;
(iv)~evaluation on procedurally generated grids with guaranteed color
diversity and on continuous state spaces with neural reward
parameterizations \cite{wulfmeier2015deepirl}; and (v)~application to
real demonstration datasets such as driving trajectories or robotic
manipulation logs.

\end{document}